\documentclass{article} 

\usepackage[scr=boondoxo,scrscaled=1.05]{mathalfa}
\usepackage{mathtools}
\usepackage{subcaption}
\usepackage{booktabs}
\usepackage{array}
\usepackage{enumitem}
\usepackage{setspace}
\usepackage[margin=2cm]{geometry}
\usepackage{newtxtext}
\usepackage{newtxmath}

\usepackage[absolute]{textpos}
\usepackage[table]{xcolor}

\newcolumntype{L}{>{$}l<{$}} % text mode "l" in an "array"
\DeclareMathAlphabet\euscr{U}{eus}{m}{n}

\makeatletter
\newcommand*{\defeq}{\mathrel{\rlap{%
			\raisebox{0.3ex}{$\m@th\cdot$}}%
		\raisebox{-0.3ex}{$\m@th\cdot$}}%
	=}
\makeatother
\begin{document}
	\title{Self-Organizing Nest Migration Dynamics Synthesis for Ant Colony Systems}
	\author{Matin~Macktoobian\footnote{matin.macktoobian@ualberta.ca}}%
	\date{Electrical and Computer Engineering Department\\ University of Alberta\\Edmonton, AB, Canada}
\maketitle

\begin{textblock}{14}(4,1)
	\noindent\textbf{\color{red}Published in ``Natural Computing'' DOI: 10.1007/s11047-022-09923-0}
\end{textblock}

\begin{abstract}
In this study, we synthesize a novel dynamical approach for ant colonies enabling them to migrate to new nest sites in a self-organizing fashion. In other words, we realize ant colony migration as a self-organizing phenotype-level collective behavior. For this purpose, we first segment the edges of the graph of ants' pathways. Then, each segment, attributed to its own pheromone profile, may host an ant. So, multiple ants may occupy an edge at the same time. Thanks to this segment-wise edge formulation, ants have more selection options in the course of their pathway determination, thereby increasing the diversity of their colony's emergent behaviors. In light of the continuous pheromone dynamics of segments, each edge owns a spatio-temporal piece-wise continuous pheromone profile in which both deposit and evaporation processes are unified. The passive dynamics of the proposed migration mechanism is sufficiently rich so that an ant colony can migrate to the vicinity of a new nest site in a self-organizing manner without any external supervision. In particular, we perform extensive simulations to test our migration dynamics applied to a colony including 500 ants traversing a pathway graph comprising 200 nodes and 4000 edges which are segmented based on various resolutions. The obtained results exhibit the effectiveness of our strategy.
\end{abstract}

\textbf{keywords}: Self-organization, Ant colony systems, Migration

%%% The next command prints the information defined in the preamble.

%%%%%%%%%%%%%%%%%%%%%%%%%%%%%%%%%%%%%%%%%%%%%%%%%%%%%%%%%%%%%%%%%%%%%%%%
\doublespacing
\section{Introduction}
Nature has always been an inspiring source for the development of efficient computational tools and algorithms based on natural complex multi-agent systems including ants \cite{dorigo1997ant}, fishes \cite{pitcher2001fish}, bees \cite{karaboga2010artificial}, birds \cite{meng2016new}, etc. These systems are attributed to particular methods based on which their agents interact with each other. Such interactions may be based on either direct communications among population members, e.g., in the case of wolf packs, or agent-environment-agent pathways, similarly to what has been observed in ant and bee swarms. The presence of environment in the latter form of communications between these natural complex systems often embarks on the emergence of useful ideas in the course of replicating these natural systems for engineering purposes. In particular, pheromone-based communications of ants \cite{kydonieus2019insect} have been the subject of numerous research successfully yielding efficient algorithms useful for solving various problems in the scope of optimization \cite{al2017max}, power scheduling \cite{abbasy2007ant}, network routing \cite{li2013adaptive} and so on. It is worth nothing that relatively simple genotypes of ants and the pheromone dynamics of their trails induce information-driven rich phenotypes. Moreover, the success of an ant-colony-based algorithm mostly requires a significant similarity between the elements of its target problem statement and the food-searching behavior of ants. However, there are classes of collective behaviors whose level of complexity may not be properly handled by the classical dynamics of ant colony systems. Self-organizing nest migration is a major representative of this group.   

Self-organization is the study of macro behaviors of (natural or artificial) swarm systems emerged from the micro behaviors and interactions of their constituent agents \cite{camazine2020self}. A unique feature of self-organization is that natural swarms' multi-modal behaviors are passively conducted based on the inherent dynamics of these systems. So, there is no central or even distributed controller to orchestrate their phenotypes. This feature makes the artificial realization of non-trivial self-organization extremely challenging for engineering purposes. The most prevalent self-organizing behavior of ant colony systems is their food-searching process in which ants lay down pheromone molecules while traversing their paths. The pheromone density of each path then determines the likelihood of its selection by other ants. Instead, The evaporation of pheromone molecules counteracts the entropy-decreasing effect of deposited pheromones. Thus, the whole colony eventually converges to an optimal path from their nest to the location of a food source. Given the magnificent effectiveness of the mechanism above to naturally conduct its desired behavior, ant colonies may potentially be capable of emerging more complex and diverse self-organizing behaviors. However, in the course of their computational modeling, some common assumptions, including the following cases, often restrict their potentials in the cited regard.   
\begin{enumerate}
	\item \textit{Simplicity of dynamics and discreteness of interaction models}: step-wise discrete deposit and evaporation dynamics of pheromone molecules may be changed to (piece-wise) continuous evolutions. Thus, one expects to achieve models which manifest more diverse behavioral traits.
	\item \textit{Shallow and relatively sparse interactions of ants}: In a classical point of view, each edge of a pathway graph associated with a colony may only be occupied by a single ant. But, one may increase the edge resolution to let more than one ant change the pheromone profile of each edge. As a result, the distribution of pheromone on an edge becomes a spatio-temporal signal which may potentially express complicated patterns of pheromone. Those pheromone patterns, thus, are promising candidates to exhibit new behavioral trends at macro levels.
\end{enumerate}
The relaxation of such assumptions may elevate the richness of the swarm dynamics of ant colonies, thereby paving the way for them to emerge more self-organizing modes of behavior. Thus, in a bio-inspired engineering perspective associated with the control theory, self-organization may effectively move the control burden from a controller's side to the system's natural dynamics. In other words, passivity of a self-organizing system may be exploited to activate its various behavioral modes instead of designing complex controllers to make the system exhibit those behaviors. These generalizations may indeed provide insights to establish more versatile ant-colony-based algorithms to be of service to many applications of self-organization such as, multi-robot systems \cite{macktoobian2017optimal}, wireless sensor networks \cite{biswas2006self}, and so forth.

Migration is a complicated collective macro behavior in pheromone-based systems including ant colonies. Natural ant colonies frequently migrate to new nest sites \cite{o2016migration} because of a variety of reasons the most major of which is severe eco-system alterations, stemmed from, e.g., harsh oil or chemical contaminations, fire, to name a few \cite{leclerc2018impact,skaldina2018ants}. If the food resources in a reasonable distance from a colony's nest become scarce, the colony's population has to put a lot of effort into gathering food from farther areas. Alternatively, in the presence of a noticeably prolific area abounding with food resources, the migration of the colony to the vicinity of the new area overall improves the nutritional state of the colony. So, the nest may totally be migrated to the vicinity of the discovered food source. We stress that migration can be a means of resilience and adaptation which may be exploited to engineer more efficient computational and algorithmic strategies based on such synergies.

Migration seeking in ant-colony-based algorithms is also notably motivated by different engineering applications. For example, node deployment in wireless sensor networks may be realized by the movement of a node from one site to another \cite{ashraf2018multi}, that is, the re-organization of the colony associated with the node and its peers. Another strategy to network migration by ant colony algorithms is the direct optimization of all interactions from one configuration to another one, as reported in \cite{turk2010network}. However, this scheme inadequately lacks scalability which is an essential factor to effectively adapt swarm-intelligence-based methods for real-world problems.

In this article, we propose a novel self-organizing model of pheromone dynamics for ant colonies according to which a colony gradually migrates from its nest site to another location. This migration dynamics is embedded into the pheromone evolution dynamics of pathways of ants, thereby requiring no external supervision, i.e., replicating a collective self-organizing behavior. To achieve the quoted objective, the major contribution of ours is four-fold as follows.
\begin{enumerate}[label=\arabic*.]
	\item In classical ant colony models, the pheromone levels of all points on each edge are identical to each other. However, this assumption is not realistic because the sooner some pheromone molecules are deposited, the more percentage of them is evaporated compared to another amount of pheromone deposited at a later time. To realize this more realistic scenario, we divide each edge to a set of segments. The pheromone profile of each segment is an exponential, thus continuous, signal. So, the overall pheromone profile of an edge is a piece-wise integration of those of its segments.
	\item We increase the freedom of ants in the course of moving from one edge to another by letting them randomly choose a segment of an edge which is not necessarily the first segment of that edge. Put differently, given a small radius with respect to a node, an ant residing at that node may reach the segments in that scope, even if those segments are not the immediate one connected to the node. This randomization expands the diversity of ants motions more than before. Consequently, one may observe more diverse phenotypic behaviors in the course of a colony's evolutions. 
	\item Our spatio-temporal dynamical model of pheromone profiles accurately incorporates both components of a pheromone updating process, say, deposit and evaporation, at the same time. So, one does not need separate rules for these processes, as is the case for classical ant-colony-based computational models and algorithms. 
	\item In contrast to the common ant colony models, we let the velocity of each ant slightly fluctuate in a random but stable (bounded) manner. This consideration is another source of diversity to be provided with our formulation. More importantly, as it is explained in Section \ref{subsec:beh}, constant identical velocities for all ants may activate some undesirable equilibriums of a colony, in the course of its migration process, that may give rise to the unreachability of a desired convergence profile.      
\end{enumerate}
This article is organized as follows. Section \ref{sec:methods} constitutes the formalism of our migration dynamics. Namely, we explain the edge segmentation step that is followed by the piece-wise continuous formulation of the pheromone profile of an edge. We then present a procedure animating our self-organizing colony migration. This section also comprises some behavioral implications regarding the migration dynamics. In Section \ref{sec:results}, we illustrate the results of our conducted simulations signifying the effectiveness of our self-organizing migration strategy. We share the potential streams of research based on the results of this article, as well as our concluding remarks.
\section{Methods}
\label{sec:methods}
\subsection{Edge segmentation}
\begin{figure}
	\centering\includegraphics[scale=1.5]{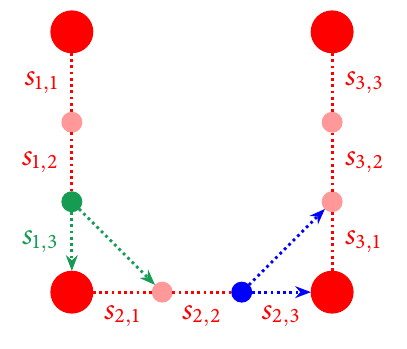}
	\caption{A schematic of edge segmentation. ($s_{i,j}$ denotes $j$th segment of $i$th edge. Large red circles represent nodes. The green and blue arrows illustrate the probabilistic transitions which may be initiated from (resp., ended at) a segment endpoint that is necessarily not a node.)}
	\label{fig:edge_segmentation}
\end{figure}

The common approach to modeling pathway graphs in ant colony systems supposes a single monolithic edge between each pair of nodes. However, as already stated, we divide each edge to a set of successive segments, so that each segment possesses its own piece of pheromone profile, as depicted in Fig. \ref{fig:edge_segmentation}. An advantage of this strategy is that a finer trajectory space entails more diverse pathway selection options, compared to the classic monolithic graphs that can only represent static trajectories. These additional degrees of dynamism in segment selection indeed increase the behavioral diversity of a colony's population.

Let $\rho_{s}$ be the pheromone profile of segment $s$ of an edge visited by an ant. Attach a clock $t$ to the segment. Then, the deposit-evaporation dynamics of the pheromone profile of $s$ is formalized by the following \textit{exponential harmonics}, say,  
\begin{equation}
\rho_{s}(t) \defeq \rho^{\star} \exp(-t),
\end{equation}  
where $\rho^{\star}$ represents the initial amount of deposited pheromone on the segment once the segment is visited by the ant. Any time an ant visits this segment, a new exponential harmonics is added to the pheromone profile of the segment according to its visit time. In particular, assume $n$ visits to $s$ associated with time slots $\{ t_{1}, \cdots, t_n\}$. Then, given the temporal Heaviside step function $\mathscr{H}_{t}(\cdot)$ defined as
\begin{equation}
\label{eq:temporal_heaviside}
\mathscr{H}_{t}(t-t_{0}) \defeq
\left\{ \begin{array}{@{\kern2.5pt}lL}
\hfill 1 & if $t \ge t_{0}$\\
\hfill 0 & if $t < t_{0}$
\end{array}\right.,
\end{equation}
the total pheromone profile of segment $s$ after $n$ visits of ants reads as follows.
\begin{equation}
\label{eq:segment_profile}
\rho_{s}(t) = \rho^{\star}\exp(-\frac{t}{\delta})\sum_{i=1}^{n}\mathscr{H}_{t}(t - t_{i})
\end{equation}
Here, $\delta > 0$ regulates the decay rate of pheromone. Equation (\ref{eq:segment_profile}) indeed represents the pheromone dynamics of segment $s$ as a progressive function of the time intervals between which ants visit this segment. Thanks to the continuous representation of the exponential term of (\ref{eq:segment_profile}), deposit and evaporation constituents of the pheromone dynamics are taken into account in a single expression, contrary to the classic discrete formulation of the dynamics in which one has to consider a specific computational term for each one of those components.
\subsection{Piece-wise continuous pheromone integration}
In classically-defined ant colony systems, no more than a single ant is allowed to occupy a particular edge at the same time. This assumption severely restricts the emergent potential dynamics of a colony compared to the relaxed scenario in that an ant may visit an edge while its other peers are also traversing it. To realize this dynamics, we employ the segment definition above to construct each edge as a series of segments. Thus, an edge may host as many ants as its number of segments as long as each ant exclusively occupies only one segment. One has to take into account the order of the segments in an edge. We apply this spatial (structural) condition to (\ref{eq:segment_profile}) throughout integrating the pheromone profiles of the segments of an edge. Assume that edge $e$ comprises in-order segment set $\{s_{1}, \cdots, s_{n}\}$ whose corresponding segment-wise pheromone profiles are $\{\rho_{s_{1}}(t), \cdots, \rho_{s_{n}}(t)\}$. Then, the pheromone dynamics of $e$ may be written as follows
\begin{equation}
\label{eq:eq}
\rho_{e}(x,t) = \sum\limits_{i=1}^{n} \rho_{s_{i}}(t)\bigg[\mathscr{H}_{x}(x_{s_{i}}) - \mathscr{H}_{x}(x - x_{s_{i+1}})\bigg],
\end{equation}
where $x_{s_{i}}$ is the coordinate of segment $s_i$ and $\mathscr{H}_{x}(\cdot)$ is the spatial Heaviside step function defined as follows. 
\begin{equation}
\label{eq:spatial_heaviside}
\mathscr{H}_{x}(x-x_{0}) \defeq
\left\{ \begin{array}{@{\kern2.5pt}lL}
\hfill 1 & if $x \ge x_{0}$\\
\hfill 0 & if $x < x_{0}$
\end{array}\right.
\end{equation}
Substituting equation (\ref{eq:segment_profile}) into equation (\ref{eq:eq}) yields the desired spatio-temporal pheromone dynamics of $e$ which reads as follows.
\begin{equation}
\rho_e(x,t) = \rho^{\star}\hspace*{-3mm}\sum\limits_{\forall \arg \{s_{i} \in e\}}\Biggl\{\biggl[\exp(-\frac{t}{\delta})\mathscr{H}_{t}(t - t_{i})\biggr]
\biggl[\mathscr{H}_{x}(x - x_{i}) - \mathscr{H}_{x}(x - x_{i+1})\biggr]\Biggr\}\label{eq:edge_profile_exact}
\end{equation}
For analytical purposes, smooth approximations of the Heaviside step functions (\ref{eq:temporal_heaviside}) and (\ref{eq:spatial_heaviside}) are more favorable. Thus, the following logistic functions
\begin{equation}
\hat{\mathscr{H}}_{t}(t-t_{0}) \defeq \frac{1}{1 + \exp\big(\!-k_{t}(t-t_{0})\big)}, 
\end{equation}
\begin{figure}
	\centering\includegraphics[scale=1.2]{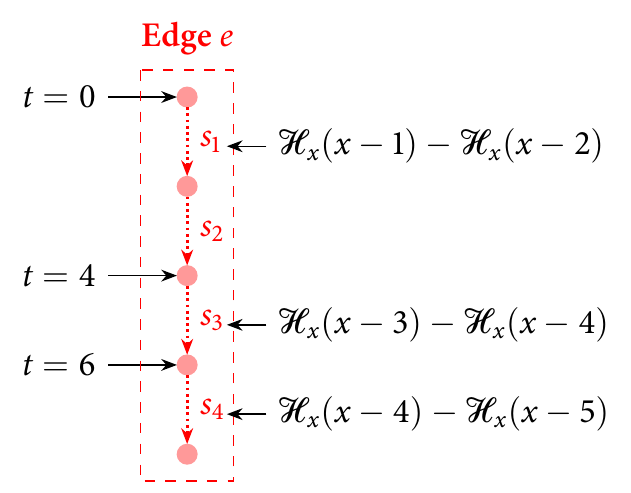}
	\caption{A hypothetical edge to represent typical conductance of pheromone excitations by some ants. (Here, threes ants start traversing segments $s_{1}$, $s_{3}$, and $s_{4}$, at times $t=0$, $t=4$, and $t=6$, respectively.)}
	\label{fig:snapshop_graph}
\end{figure}
\begin{equation}
\hat{\mathscr{H}}_{x}(x-x_{0}) \defeq \frac{1}{1 + \exp\big(\!-k_{x}(x-x_{0})\big)}, 
\end{equation}  
turn equation (\ref{eq:edge_profile_exact}) to the approximate form below.
\begin{equation}
\hat{\rho}_e(x,t) = \rho^{\star}\hspace*{-5mm}\sum\limits_{\forall \arg \{s_{i} \in e\}}\hspace*{-1mm}\Biggl\{\biggl[\exp(-\frac{t}{\delta})\hat{\mathscr{H}}_{t}(t - t_{i})\biggr]
\biggl[\hat{\mathscr{H}}_{x}(x - x_{i}) - \hat{\mathscr{H}}_{x}(x - x_{i+1})\biggr]\Biggr\}\label{eq:edge_profile_approx}
\end{equation}
So, the overall dynamics of the whole colony maybe written as follows.
\begin{figure}[t]
	\centering\includegraphics[scale=0.6]{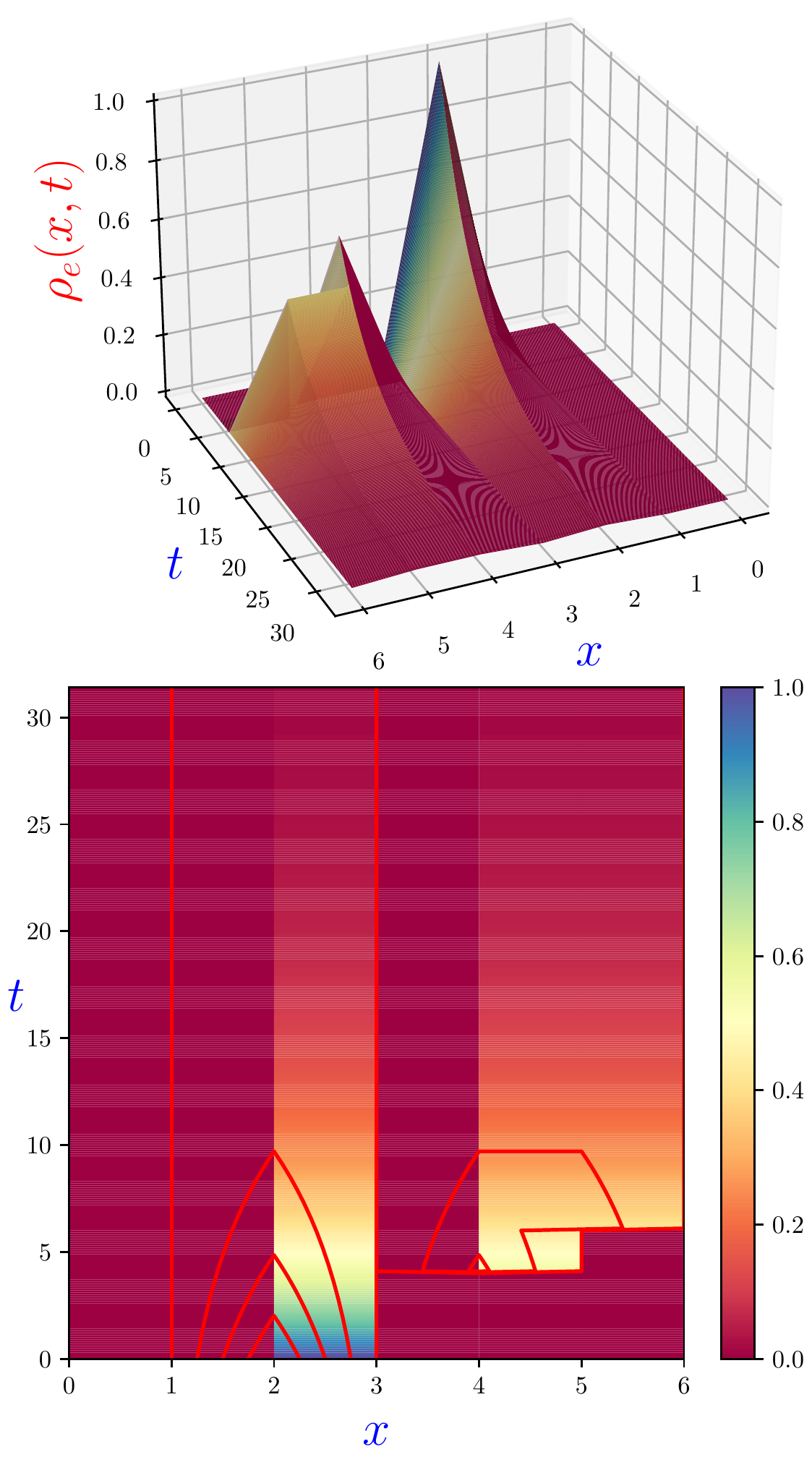}
	\caption{The precise pheromone dynamics of the hypothetical edge illustrated in figure (\ref{fig:snapshop_graph}).}
	\label{fig:EPP_snapshop}
\end{figure}
\begin{equation}
\hat{\rho}(x,t) = \rho^{\star}\sum\limits_{\forall e}\Biggl[\sum\limits_{\forall \arg \{s_{i} \in e\}}\hspace*{-1mm}\Biggl\{\biggl[\exp(-\frac{t}{\delta})\hat{\mathscr{H}}_{t}(t - t_{i})\biggr]
\biggl[\hat{\mathscr{H}}_{x}(x - x_{i}) - \hat{\mathscr{H}}_{x}(x - x_{i+1})\biggr]\Biggr\}\Biggr]\label{eq:total_profile}
\end{equation}
Figure \ref{fig:snapshop_graph} renders a simple example of an edge comprising four segments. Here, we assume that each ant stops once it reaches the immediate node on its pathway. In particular, three ants start to move on some specific segments of the edge at particular time slots. Accordingly, Fig. \ref{fig:EPP_snapshop} (resp., Fig. \ref{fig:APP_snapshop}) depicts the precise (resp., approximate) profile of the edge obtained from (\ref{eq:edge_profile_exact}) (resp., (\ref{eq:edge_profile_approx})). One notes a sufficient level of similarity between these precise and approximate versions. The gradient of the approximate pheromone profile of this example is also demonstrated in Fig. \ref{fig:grad_approax}.
\begin{figure}
	\centering\includegraphics[scale=0.6]{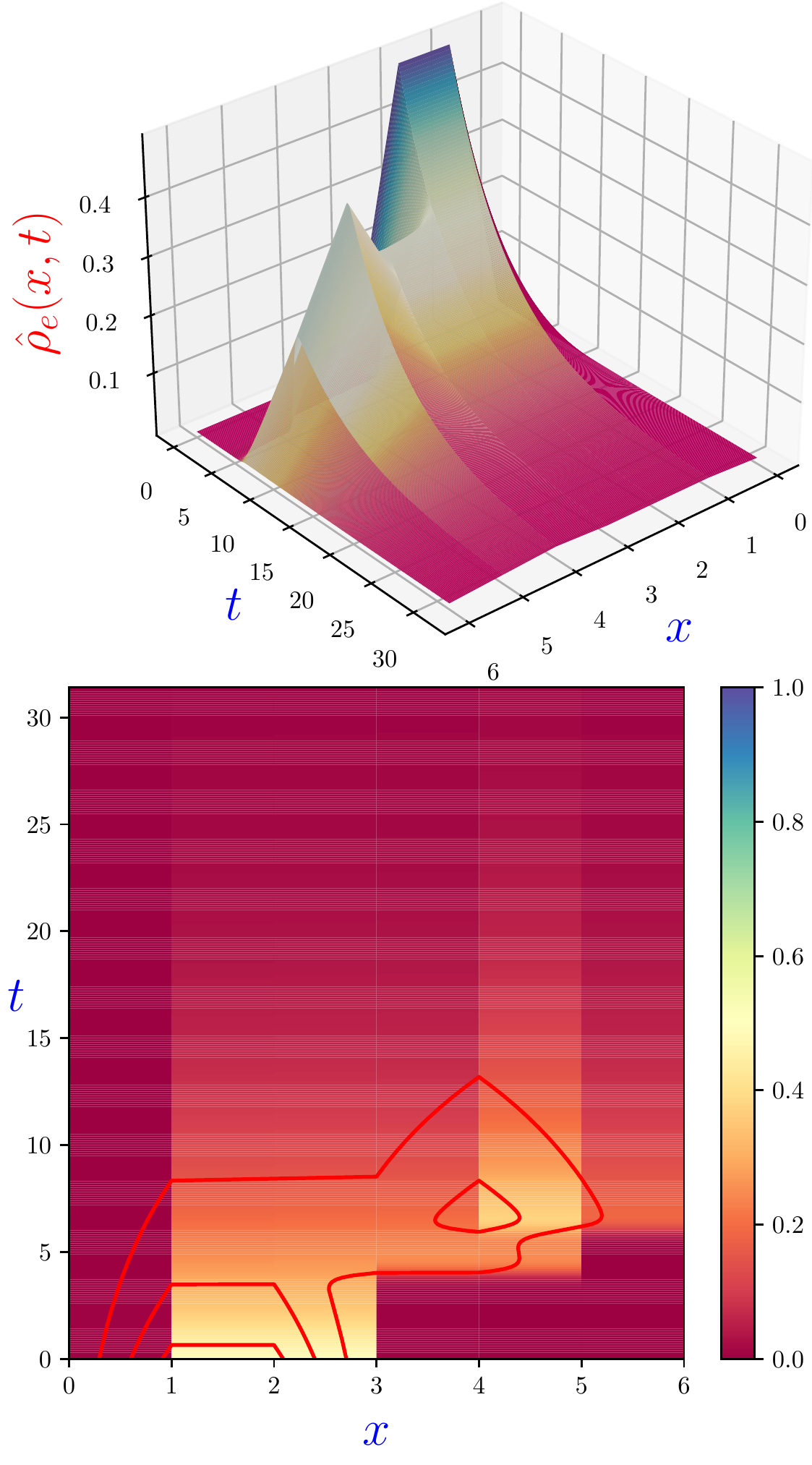}
	\caption{The approximate pheromone dynamics of a hypothetical edge illustrated in Fig. \ref{fig:snapshop_graph}.}
	\label{fig:APP_snapshop}
\end{figure}

\begin{figure}[t]
	\centering\includegraphics[scale=0.5]{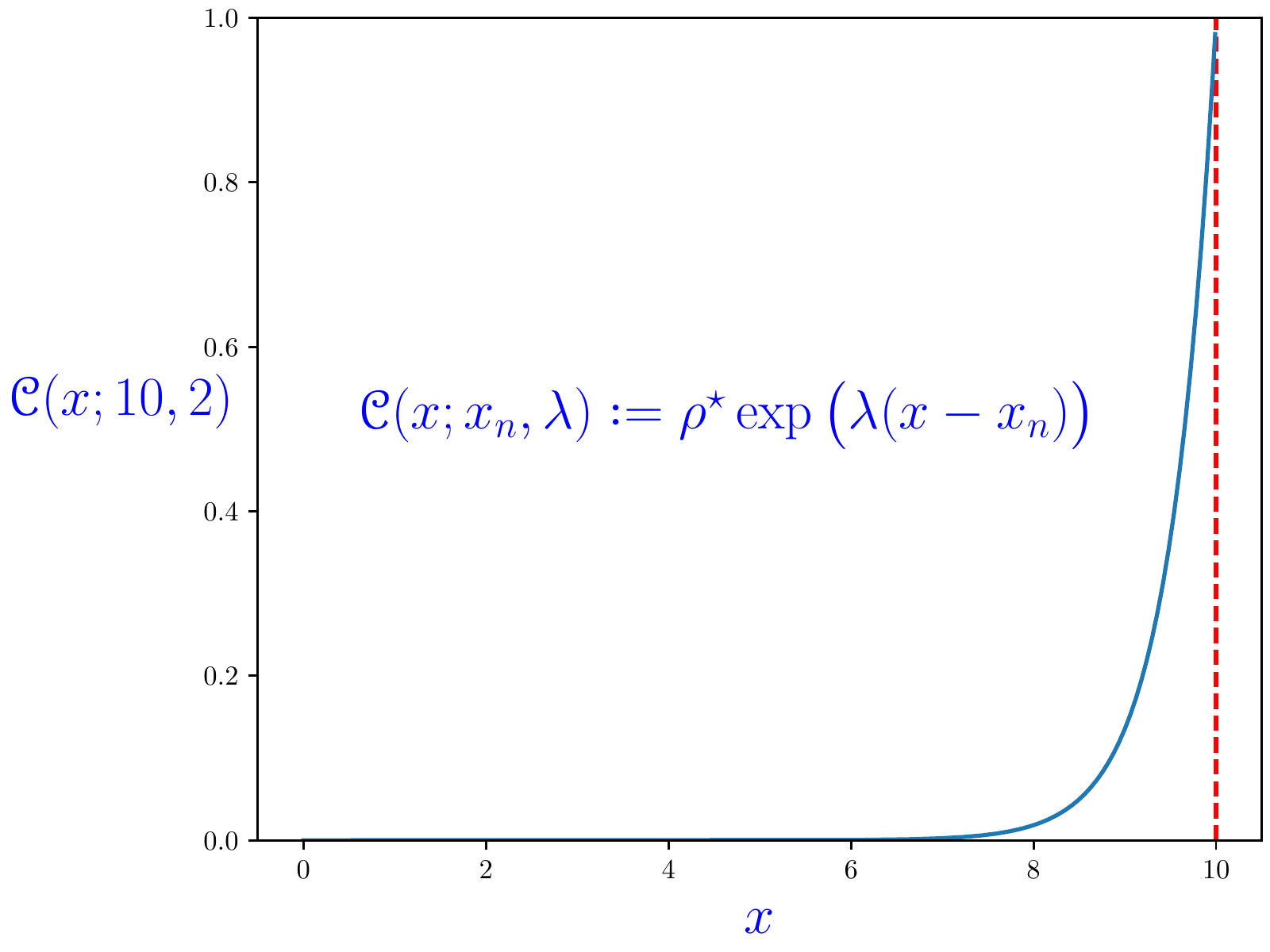}
	\caption{A typical population distribution around a hypothetical target nest site at $x=x_{n}=10$.}
	\label{fig:target_profile}
\end{figure}

A migration scenario can be accomplished if the ants of a colony eventually get absorbed toward the vicinity of a target nest site. Thus, we need to establish a convergence profile, so-called \textit{radial convergence distribution function} whose magnitude is exponentially large around a target node $x_n$ and decaying fast once an ant moves away from that point. A candidate function can be
\begin{equation}
\label{eq:radial}
\euscr{C}(x; \sigma, \lambda) \defeq \exp(\lambda x - \sigma),
\end{equation} 
where $\sigma$ and $\lambda$ are convergence radius and a scale factor, respectively, and $x_{n}$ is the location of the target nest site. Given a fixed initial deposit pheromone $\rho^{\star}$, the function shall satisfy the following boundary condition
\begin{equation}
\euscr{C}(x_{n}; \sigma, \lambda) = \exp(\lambda x_{n} - \sigma) = \rho^{\star},
\end{equation}
according to which (\ref{eq:radial}) may be written as
\begin{equation}
\euscr{C}(x; x_{n}, \lambda) = \rho^{\star}\exp\big(\lambda (x - x_{n})\big).
\end{equation}
The dynamics of a typical instance of this function is depicted in Fig. \ref{fig:target_profile}. In the sequel, the fulfillment of the following steady-state condition guarantees the convergence of the migration dynamics.
\begin{equation}
\lim\limits_{t \to \infty} \hat{\rho}_e(x,t) = \euscr{C}(x;x_{n},\lambda)
\end{equation}
\subsection{Migration mechanism}
\begin{figure}
	\centering\includegraphics[scale=0.94]{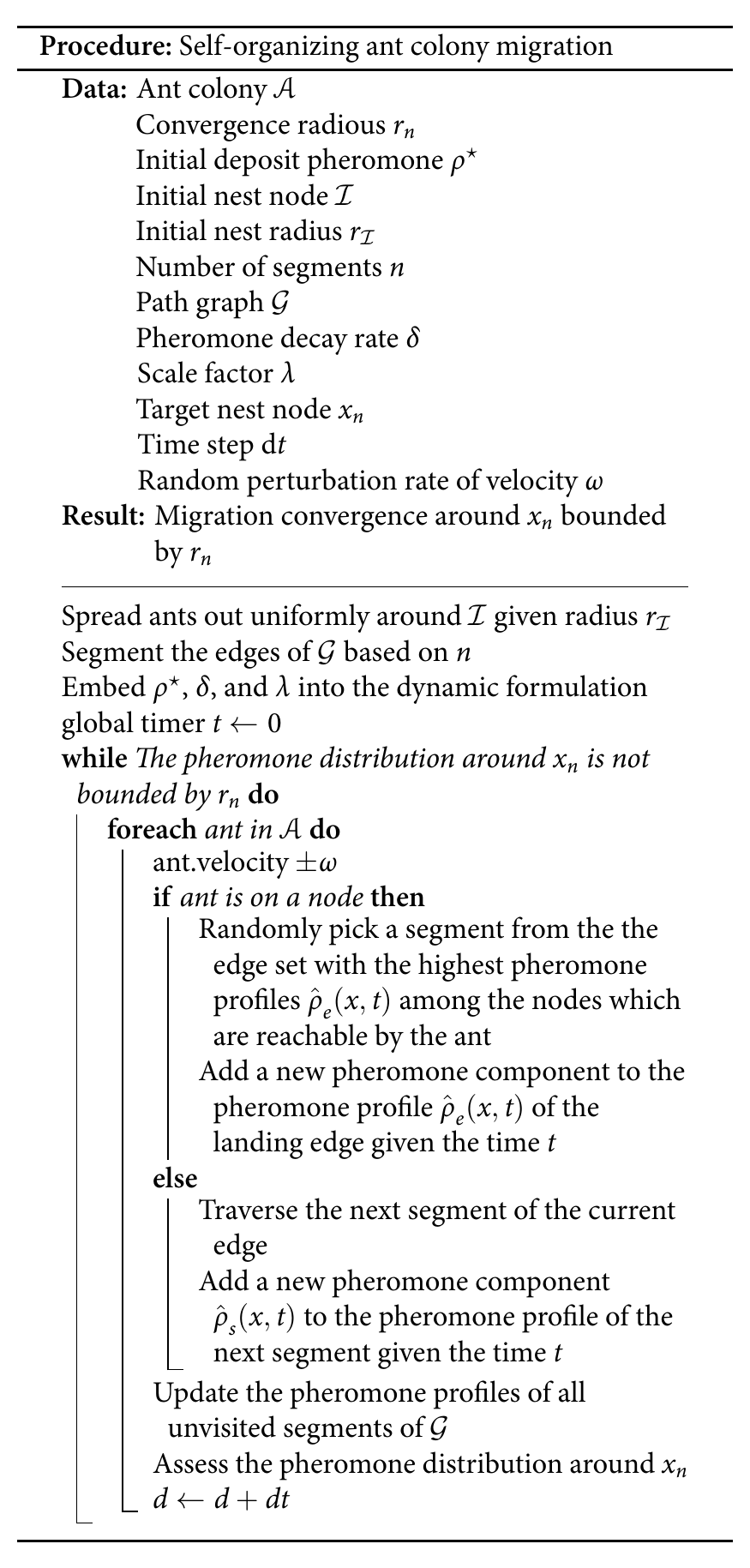}
	\caption{Self-organizing ant colony migration procedure.}
	\label{fig:alg}
\end{figure}

Our migration strategy cannot be incorporated to the classical ant colony algorithms because we allow both edge segmentation and piece-wise continuous pheromone profiles which are not applicable to those frameworks. Alternatively, figure (\ref{fig:alg}) represents a procedure according to which one can animate a colony of ants that eventually migrates from an initial nest site to a new one.

This procedure first spread all ants uniformly around the initial nest node $\mathscr{I}$ with respect to a particular radius $r_{\mathscr{I}}$. Then, all edges are segmented based on the value of the hyperparameter $n$. We also initialize a timer $t$ to globally keep track of the time slots associated with ants and their motions. At each step of the clock, we slightly perturb each ant's velocity in a random fashion. Then, ants start to traverse the pathway graph of the migration scenario based on the following rules. If an ant is at a node, it randomly picks a segment of the set of the reachable segments with the highest pheromone profiles. Then, a new exponential spike of pheromone is added to the profile of the selected segment right at the time the ant starts to traverse it. If the ant is not at a node, it is essentially somewhere in the exclusive scope of its current edge. So, it can only pick the next segment which is preceded by its current segment. Similarly to the previous case, the pheromone profile of the next segment is updated once the ant reaches it. Finally, the decay element of the piece-wise continuous pheromone profiles are also taken into account for all the segments which were not visited by any ant. This procedure continues until the ant (or equivalently, pheromone) distribution around the target nest site sufficiently resembles that of the desired one.       
\begin{figure}[t]
	\centering\includegraphics[scale=0.3]{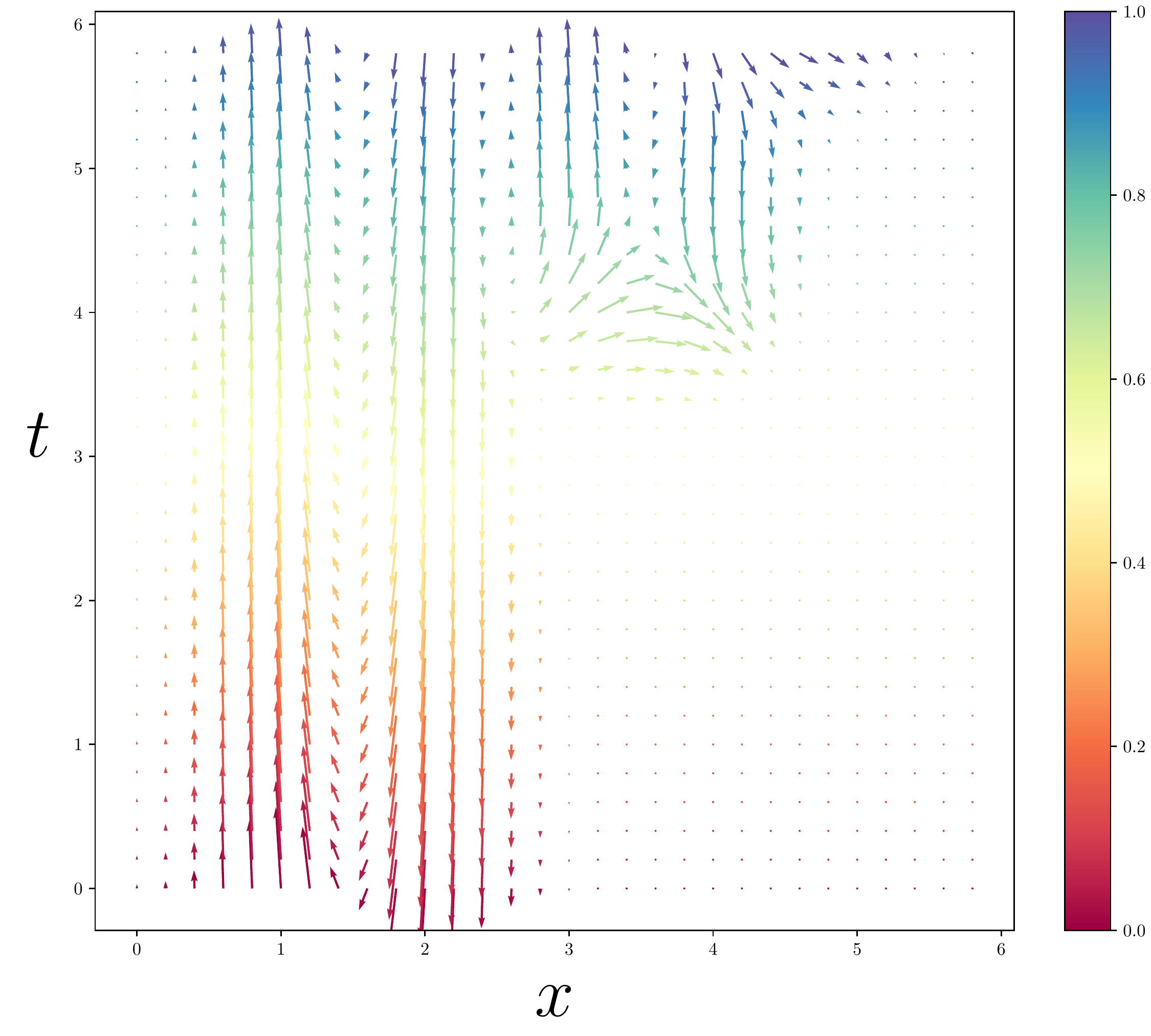}
	\caption{The gradient of the approximate pheromone profile of the edge depicted in Fig. \ref{fig:snapshop_graph}.}
	\label{fig:grad_approax}
\end{figure}

The hyperparameters setting of the described procedure is discussed in Section \ref{sec:results}.
\subsection{Behavioral implications of the migration formalism}
\label{subsec:beh}
In this section, we supply insights into some behavioral limits of a colony under the dynamical framework discussed in the previous sections. In particular, we study the existence of both temporal and spatial equilibriums of a migration process orchestrated by our proposed method. In this regard, we derive the necessary conditions for those existence and interpret their implications. 
\subsubsection{Temporal equilibriums}
We seek the condition based on which a migrating colony can have a temporal equilibrium. A temporal equilibrium is the one that, once is reached, does not vary over time. Formally speaking, the condition implies that
\begin{equation}
\nabla\!_{t} \hat{\rho}(x,t) = 0.
\end{equation}
Feeding (\ref{eq:total_profile}) into the equation above, after some simplifications, yields the following condition,
\begin{equation}
\sum\limits_{\forall e}\sum\limits_{\forall \arg \{s_{i} \in e\}}\Bigl[	
\hat{\mathscr{H}}_{t}(t-t_{i})\big(1-\hat{\mathscr{H}}_{t}(t-t_{i})\big)\Bigr] = \frac{1}{\delta},
\end{equation}
which in general may have stationary solutions. However, in the case of very slow evaporation ($\delta \to \infty$), we have 
\begin{equation}
\sum\limits_{\forall e}\sum\limits_{\forall \arg \{s_{i} \in e\}}\Bigl[\hat{\mathscr{H}}_{t}(t-t_{i})\Bigr] = 1,
\end{equation}
which is not a feasible condition. We argue that this result is reasonable because pheromone dynamics without evaporation cannot be stable. So, one does not expect to reach any temporal equilibrium no matter how long a migration process may be continued.
\subsubsection{Spatial equilibriums}
In the same vein as the previous section, we are interested in investigating the existence of any spatial equilibrium associated with our migration dynamics. To be specific, the existence of a spatial equilibrium entails that, given a fixed time slot, the pheromone dynamics all over the pathway graph are stationary. So, considering
\begin{equation}
\nabla\!_{x} \hat{\rho}(x,t) = 0,
\end{equation}
and taking (\ref{eq:total_profile}) into account, we finally yield
\begin{equation}
\sum\limits_{\forall e}\sum\limits_{\forall \arg \{s_{i} \in e\}}\Biggl\{\hat{\mathscr{H}}_{x}(x-x_{i})\Bigl[1- \hat{\mathscr{H}}_{x}(x-x_{i})\Bigr] - \hat{\mathscr{H}}_{x}(x-x_{i+1})\Bigl[(1-\hat{\mathscr{H}}_{x}(x-x_{i+1}))\Bigr]\Biggr\} = 0.
\end{equation}
Since we have already assumed velocities of ants vary, the equation above has no solutions. But, if we force a constant velocity to all ants at all times, at least one of the solutions below has to hold.
\begin{eqnarray}
\left\{ \begin{array}{@{\kern2.5pt}lL}
x_{i} = x_{i+1}\\
x_{i} + x_{i+1} = 1
\end{array}\right.
\end{eqnarray}
The first solution is a trivial one because if the velocities of ants are the same, and the lengths of the edges are also identical to each other, then the whole colony is in a spatially-synchronized mode. In this mode, ants move from one node to another in exactly the same pace. Additionally, if the lengths of the edges are equal, their time-independent pheromone profiles are identical. Put differently, if temporal evolutions of the pheromone profiles halt, the distributions of the pheromone on all edges do not vary. So, ants can not perceive any spatial variation based on what they sense according to the trails of pheromone. Obviously, time cannot halt, so this does practically not happen. 

The second solution, however, is non-trivial and needs to be avoided, should one takes the assumption of constant velocities into account. Namely, this mode may interfere the evolution of a colony toward its convergence in the course of a migration process. However, since we do not enforce any equality condition on velocities in our approach, no spatial gradient does vanish. 
\section{Results}
\label{sec:results}
\subsection{Simulations setup}
We use NetworkX library \cite{hagberg2008exploring} to create the pathway graph of our simulations. The simulations are all developed in Python 3.7. We conduct
the simulations on an ASUS ZenBook UX410UAR with an Intel
Core i7-8550U at 1.8 GHz x4 processor, Intel UHD Graphics
620 graphic card on an Microsoft Windows machine. The structural configuration of the pathway graph, in addition to the remainder of the hyperparameters involved in the migration dynamics procedure, are listed in Table \ref{tbl:tbl1}. The initial nest node and the target nest nodes are not varied throughout the tests. We intentionally select the farthest nodes of the pathway graph to represent those critical nodes for the purpose of providing the most complicated migration cases in terms of both required time and algorithmic effort. 
\begin{table}
	\centering\caption{The setting of hyperparameters in performed simulations}
	\label{tbl:tbl1}
	\centering\includegraphics[scale=1]{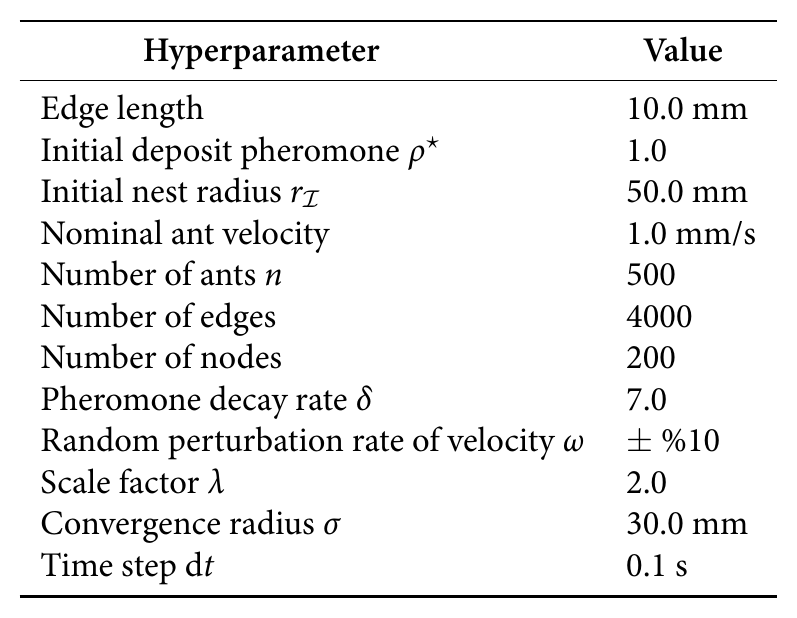}
\end{table}
\subsection{Metrics and performance report}
In addition to the fixed hyperparameters reported in Table \ref{tbl:tbl1}, one may be interested in the impact of various convergence radii and edge segmentation resolutions on the performance of the migration mechanism. The metrics to assess the functionality and the efficiency of the mechanism are $r_n$-convergence rate and convergence time. $r_n$-convergence rate determines the percentage of the ants which reside in a distance less than a particular convergence radius from a specific target nest node. In other words, $r_n$-convergence rate determines how well the migration mechanism can gather a colony around its new nest site. Convergence time also addresses the computational speed of the mechanism with respect to various convergence radii and edge segmentation resolutions. 

We take 9 simulation scenarios into account associated with a grid space of 3 convergence radii (20, 30, 40), and 3 edge segmentation resolutions (0, 5, 10). One notes that all of the planned convergence radii are less than the initial nest radius, which is 50. This setup is indeed considered to study the capabilities of the migration mechanism to condense ant colonies in the course of their migrations. In this regard, the reported $r_n$-convergence rates demonstrate the strong condensation capabilities of the migration mechanism. This feature is significantly important in a number of engineering applications such as and cooperative coordination of multi-robot systems \cite{macktoobian2022meta}.  

\begin{figure*}
	\centering
	\begin{subfigure}{.33\textwidth}
		\centering
		\includegraphics[scale=0.4]{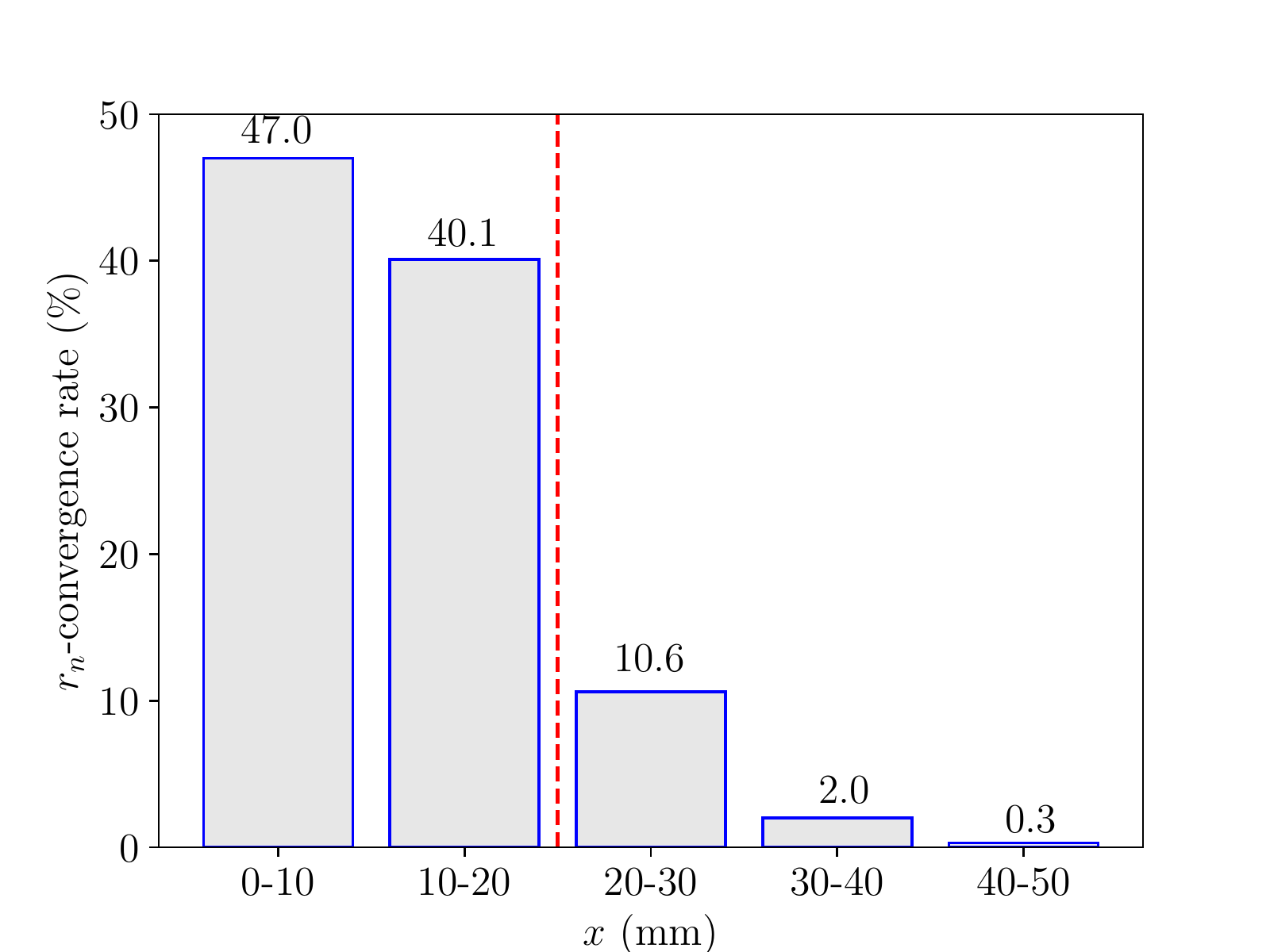}
		\caption{$r_n = 20$ mm}
		\label{fig:sub1}
	\end{subfigure}%
	\begin{subfigure}{.33\textwidth}
		\centering
		\includegraphics[scale=0.4]{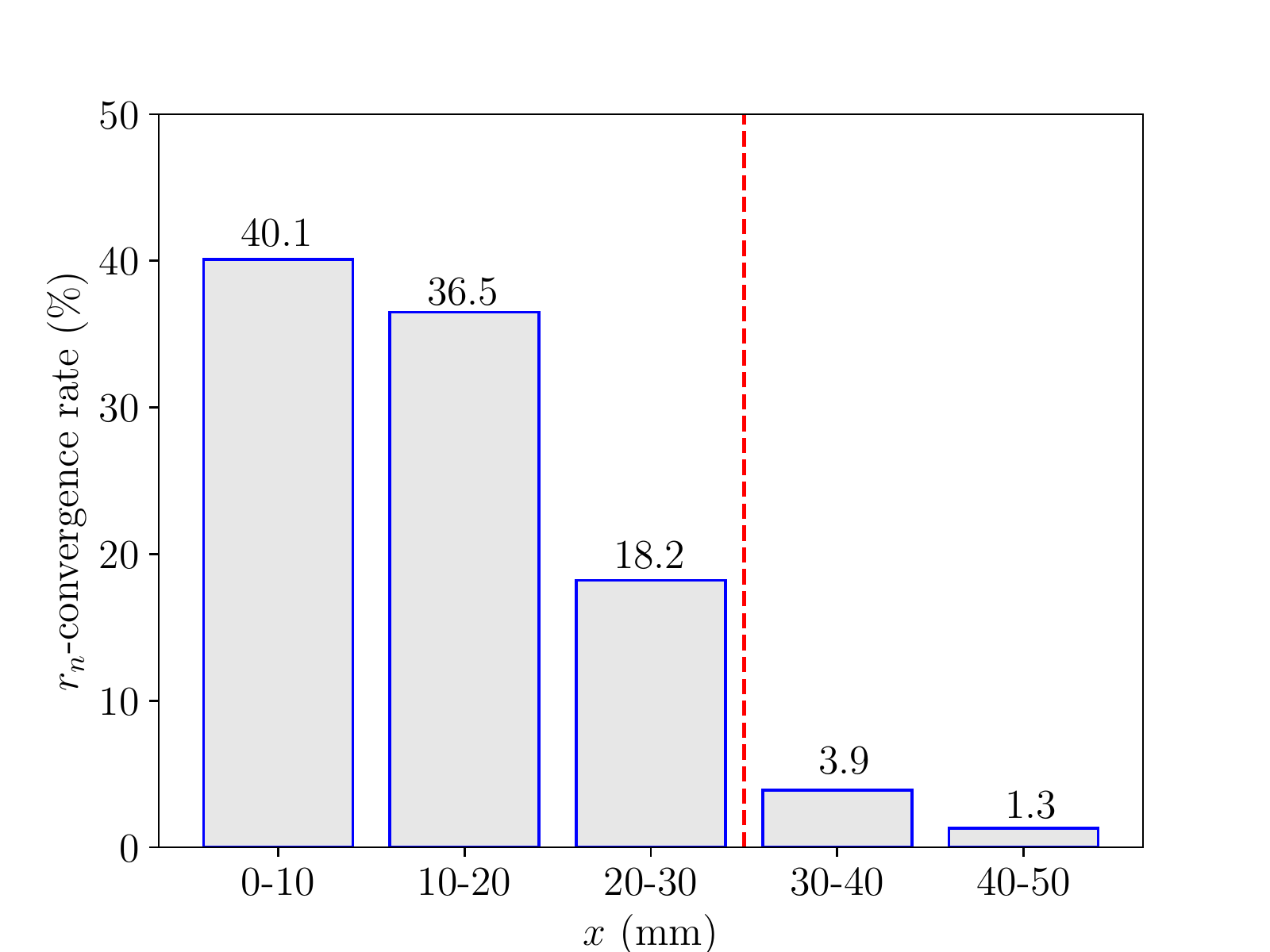}
		\caption{$r_n = 30$ mm}
		\label{fig:sub2}
	\end{subfigure}%
	\begin{subfigure}{.33\textwidth}
		\centering
		\includegraphics[scale=0.4]{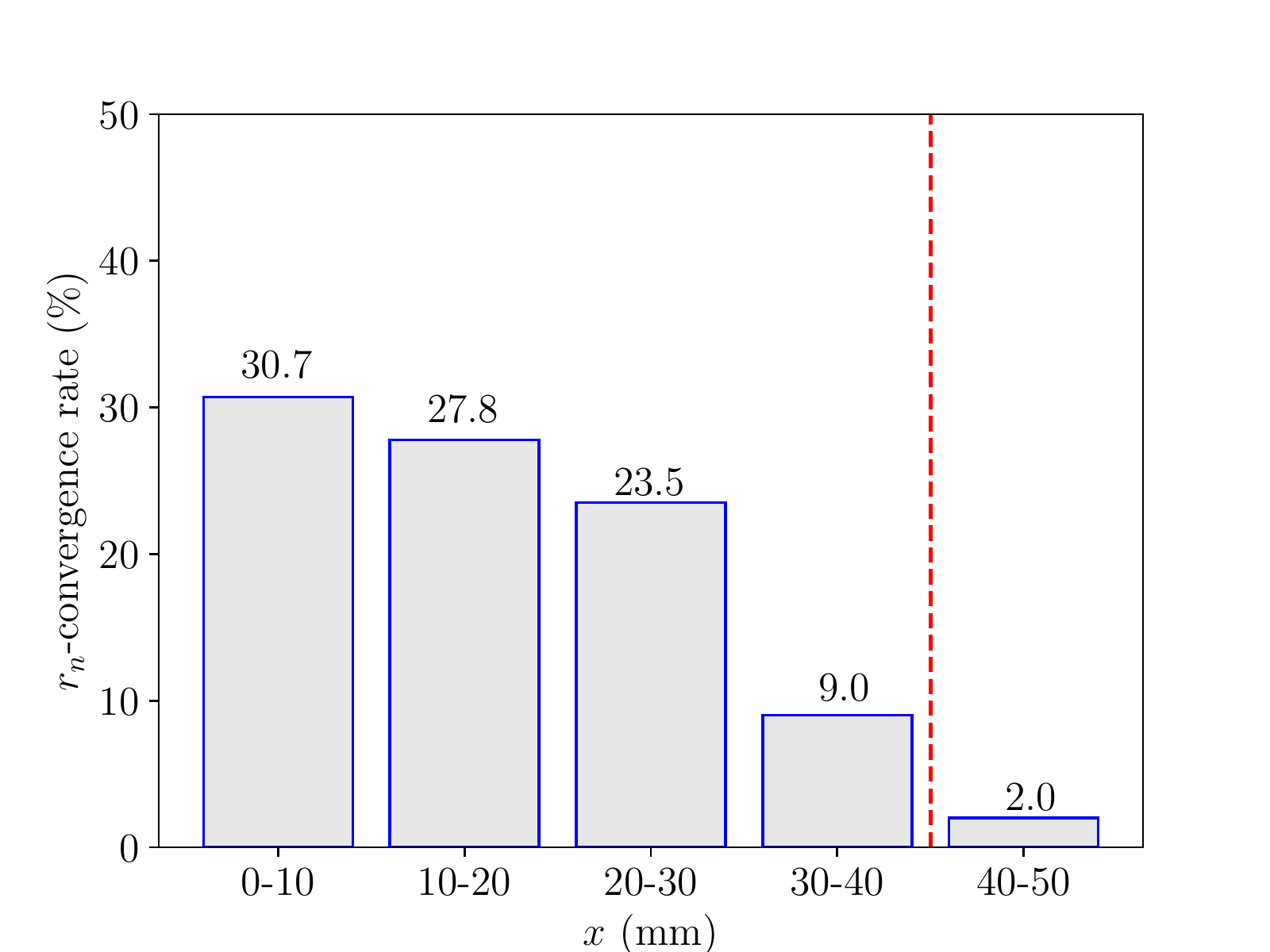}
		\caption{$r_n = 40$ mm}
		\label{fig:sub3}
	\end{subfigure}%
	\caption{The detailed distributions of ants after the migration processes converged around the target node corresponding to the highlighted cells of Table \ref{tbl:tbl}.}
	\label{fig:r_n}
\end{figure*}

As one expects, the larger a convergence radius is, the higher its corresponding $r_n$-convergence rate will be because more ants are qualified to be taken as those residing in the vicinity of the target nest node. Moreover, as we already stated, increasing edge segmentation resolution magnifies the diversity of path selections of a colony's population. Thus, given a fixed convergence radius, increasing the edge segmentation resolution overall improve the $r_n$-convergence rate. in this regard, Fig. \ref{fig:r_n} represents the detailed distributions of the ants after migration convergences associated with the highlighted cells of Table \ref{tbl:tbl}. One can observe that the migration mechanism effectively gather the majority of ants around the desired target node given various convergence radii.

Convergence time is also reciprocally proportional to both convergence radius and edge segmentation resolution. Convergence radius reduction enforces the migration mechanism to spend more time on condensing ants in smaller regions. Smaller edge segmentation resolutions also make the mechanism search for more different combinations of available edges because coarse-grained edges do not provide ants with high degrees of freedom in any pathway selection.  

\begin{figure}
	\centering\includegraphics[scale=0.35]{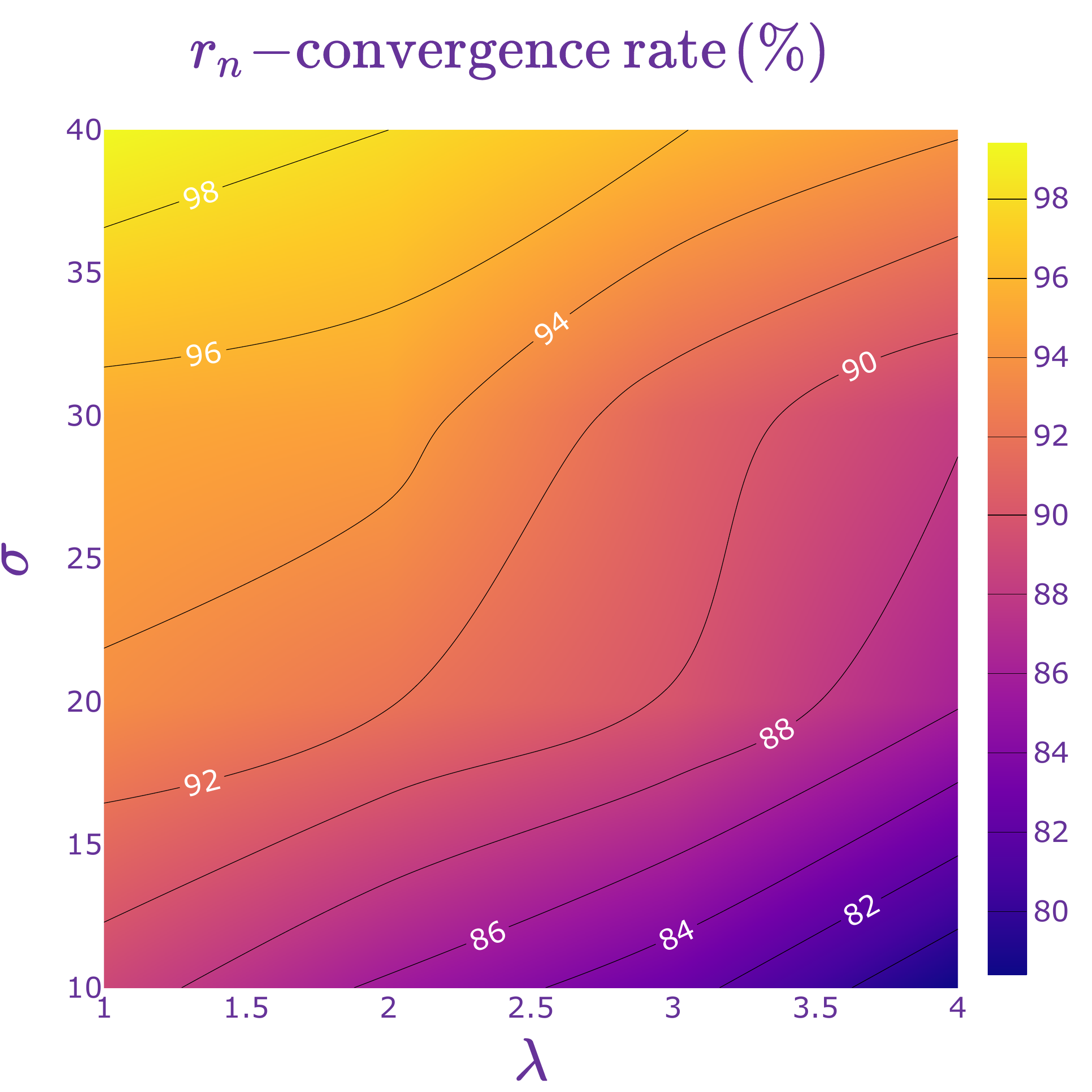}
	\caption{The dynamics of $r_{n}$-convergence rate with the respect to scale factor and convergence radius.}
	\label{fig:countor}
\end{figure}

There is also a trade-off between the compactness of a swarm in view of its $r_{n}$-convergence rate around a new site point and a desired convergence radius. Accordingly, as Fig. \ref{fig:countor} exhibits, the larger convergence radius and/or the smaller the scale factor are, the higher the $r_{n}$-convergence rate is. Here, edge segmentation resolution is set to 10. This observation makes sense because the larger the convergence radius is, the more ants contribute to the increment of its $r_{n}$-convergence rate. 

Overall, the proposed self-organizing migration dynamics demonstrates substantial abilities in collective orchestration of ant colonies without external supervision and only based on their rich passive dynamical characteristics.
\begin{table}
	\centering\caption{The results of the performed simulated self-organizing ant colony migrations}
	\label{tbl:tbl}
	\centering\includegraphics[scale=0.8]{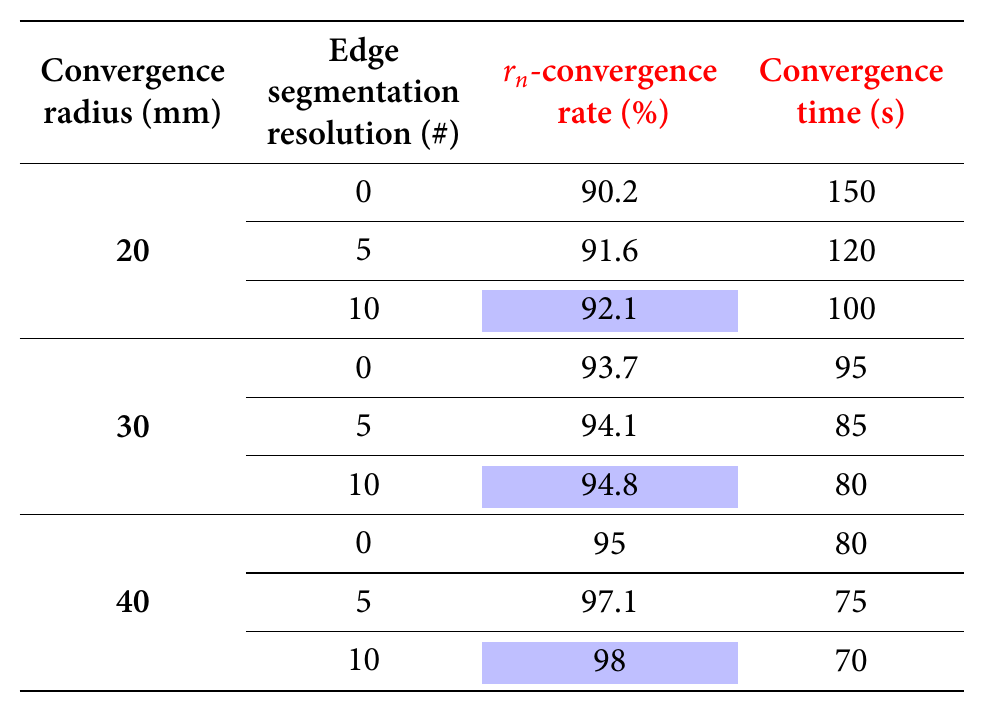}
\end{table}
\section{Conclusions}
\label{sec:conclusions}
In this study, we reformulated pheromone-based ant interactions to realize self-organizing ant colony migration. For this purpose, we first generalized the notion of edge pathways to constitute many segments each of which is able to host an ant. We then established a piece-wise continuous pheromone profile for each edge as the accumulation of those of its segments. This pheromone formulation effectively manages both deposit and evaporation elements of pheromone-based systems by only one single exponential expression. We wrapped our spatio-temporal migration strategy into a procedure so that ants can choose various trails in the course of interacting with pheromone trails. Thanks to the expressivity of the passive dynamics of our scheme, various non-trivial simulations of ant colony migration were successfully conducted in a self-organizing manner.
%The applied equilibrium analyses in this article solely investigated the spatial and temporal equilibriums. However, the found equilibriums were either infeasible to reach or undesirable due to their effects. We argue that further extensive spatio-temporal equilibrium analysis may shed light on less-known aspects of this dynamical framework. One may find behaviorally interesting equilibriums that elevate the current uni-modal phenotype model of this migration procedure to a multi-modal one.

This study may also potentially lead to the emergence of interesting venues regarding the development of self-organizing systems expressing more complex collective behaviors. For example, transferring the ideas of this work to the realm of bee colonies and their unique eco-system of genotypes may give rise to further self-organization manifestations. As another future stream of research, one may focus on the cooperativity levels of genotype-level interactions in ant colonies and revisiting the fundamental assumptions of theirs aiming for discovering new phenotype-level behaviors. As an instance, the approach of this study to cooperations between ants, similarly to the remainder of the literature, e.g., \cite{wang2021assembly,deng2019improved,deng2020effective}, is symbiotic. But the definition of confrontational sub-colonies can potentially unfold novel collective behaviors. The addition of defense mechanisms to passively coordinate colony members can be an angle to approach the quoted idea. Accordingly, one may take various colony-level safety requirements into account such as queen safety, colony territory preserving, divide-and-conquer-based force assignment to cope with multiple intruders, and so on.
%\begin{acks}
%If you wish to include any acknowledgments in your paper (e.g., to 
%people or funding agencies), please do so using the `\texttt{acks}' 
%environment. Note that the text of your acknowledgments will be omitted
%if you compile your document with the `\texttt{anonymous}' option.
%\end{acks}

%%%%%%%%%%%%%%%%%%%%%%%%%%%%%%%%%%%%%%%%%%%%%%%%%%%%%%%%%%%%%%%%%%%%%%%%

%%% The next two lines define, first, the bibliography style to be 
%%% applied, and, second, the bibliography file to be used.

\bibliographystyle{IEEEtran} 
\bibliography{references}

\end{document}